%% file: Paper_format.tex
\documentclass[conference, letterpaper]{IEEEtran}
\usepackage{arabtex}
\usepackage{utf8}

\usepackage{times}
\usepackage{url}
\usepackage{latexsym}
\usepackage{amsmath}
\usepackage{booktabs}
\usepackage{graphicx}
\usepackage{url}
\usepackage{adjustbox,multirow}
\usepackage{tabularx}
\usepackage{amsmath}
\usepackage{pgf}
\usepackage{color,soul}
\usepackage{float}
\usepackage[ruled,vlined, linesnumbered]{algorithm2e}
\usepackage{textcomp}
\usepackage{subfig}
\usepackage{float}
\usepackage{xcolor}
\usepackage{comment}
\usepackage{hyperref}
\usepackage{float}
\floatstyle{plaintop}
\restylefloat{table}

\newcommand{\floor}[1]{\left\lfloor #1 \right\rfloor}

\ifCLASSINFOpdf
\else
\fi
\usepackage{fancyhdr}

\renewcommand{\thispagestyle}[2]{}

\fancypagestyle{plain}{
        \fancyhead{}
        \fancyhead[C]{first page center header}
        \fancyfoot{}
        \fancyfoot[C]{first page center footer}
}
\begin{document}
\setcode{utf8}

%
\title{Zero-Resource Multi-Dialectal Arabic Natural Language Understanding}

\author{\IEEEauthorblockN{Muhammad Khalifa}
\IEEEauthorblockA{Cairo University, Egypt\\ m.khalifa@grad.fci-cu.edu.eg\\}
\and
\IEEEauthorblockN{Hesham Hassan}
\IEEEauthorblockA{Cairo University, Egypt\\ h.hassan@fci-cu.edu.eg}
\and
\IEEEauthorblockN{Aly Fahmy}
\IEEEauthorblockA{Cairo University, Egypt\\ aly.fahmy@cu.edu.eg }}


%


\maketitle

\begin{abstract}
A reasonable amount of annotated data is required for fine-tuning pre-trained language models (PLM) on downstream tasks. However, obtaining labeled examples for different language varieties can be costly. In this paper, we investigate the zero-shot performance on Dialectal Arabic (DA) when fine-tuning a PLM on modern standard Arabic (MSA) data only --- identifying a significant performance drop when evaluating such models on DA. To remedy such performance drop, we propose self-training with unlabeled DA data and apply it in the context of named entity recognition (NER), part-of-speech (POS) tagging, and sarcasm detection (SRD) on several DA varieties. Our results demonstrate the effectiveness of self-training with unlabeled DA data: improving zero-shot MSA-to-DA transfer by as large as $\sim$10\% F$_1$ (NER), 2\% accuracy (POS tagging), and 4.5\% F$_1$ (SRD). We conduct an ablation experiment and show that the performance boost observed directly results from the unlabeled DA examples used for self-training. Our work opens up opportunities for leveraging the relatively abundant labeled MSA datasets to develop DA models for zero and low-resource dialects. We also report new state-of-the-art performance on all three tasks and open-source our fine-tuned models for the research community. 
\end{abstract}


\begin{IEEEkeywords}
Arabic Natural Language Processing; Natural Language Understanding; Low-resource learning;
\end{IEEEkeywords}

%

\section{Introduction}
\label{intro}

%
%
    


Neural language models~\cite{xu2000can,bengio2003neural} with contextual word representations~\cite{DBLP:conf/naacl/PetersNIGCLZ18} have become dominant for a wide range of Natural Language Processing (NLP) downstream tasks. More precisely, contextual representations from transformer-based~\cite{vaswani2017attention} language models~\cite{devlin2018bert,liu2019roberta}, pre-trained on large amounts of raw data and then fine-tuned on labeled tasks-specific data, has produced state-of-the-art performance on many tasks, even when using fewer labeled examples. Such tasks include question answering \cite{yang2019enhancing}, text classification~\cite{liu2019roberta}, named entity recognition (NER), and part-of-speech (POS) tagging~\cite{tsai2019small,conneau2019unsupervised}.

Typically, such language models see a huge amount of data during pre-training, which could mistakenly lead us to assume they have a strong generalization capability even in situations where the language varieties seen at test time are different from those the language model was fine-tuned on. To investigate this particular situation, we first study the impact of using a language model pre-trained on huge Arabic corpora for two popular sequence tagging tasks (NER and POS tagging) and one text classification task (sarcasm detection) when fine-tuned on available labeled data, regardless of language variety (Section~\ref{sec:finetune}). To test the model utility for tasks based on exclusively dialectal Arabic (DA), we then remove all dialectal data from the training splits and fine-tune a model only on MSA. Evaluating such a model in a \textit{zero-shot} setting, i.e., on Egyptian (EGY), Gulf (GLF), and Levantine (LEV) varieties, we observe a significant performance drop. This shows the somewhat brittle ability of pre-trained language models without dialect-specific fine-tuning. 

Unfortunately, the scarcity of labeled DA resources covering sufficient tasks and dialectal varieties has significantly slowed down research on DA \cite{darwish2020effective}. Consequently, a question arises: ``How can we develop models nuanced to downstream tasks in dialectal contexts without annotated DA examples?". We apply self-training, a classical semi-supervised approach where we augment the training data with confidently-predicted dialectal data. We empirically show that self-training is indeed an effective strategy, which proves to be useful in \textit{zero-shot} (where no gold dialectal data are included in training set) independently as well as with self-training (Sections ~\ref{sec:msa-da-zero} and~\ref{sec:msa-da-zero-st}, respectively). 

Our experiments reveal that self-training is always a useful strategy that \textit{consistently} improves over mere fine-tuning. In order to understand why this is the case (i.e., why combining self-training with fine-tuning yields better results than mere fine-tuning), we perform an extensive error analysis based on our NER data. We discover that self-training helps the model most with improving false positives (approximately 59.7\%). This includes in cases involving DA tokens whose MSA orthographic counterparts \cite{shaalan2014survey}  are either named entities or trigger words that frequently co-occur with named entities in MSA. Interestingly, such out-of-MSA tokens occur in highly dialectal contexts (e.g., interjections and idiomatic expressions employed in interpersonal social media communication) or ones where the social media context in which the language (DA) employed affords more freedom of speech and a platform for political satire. We present our error analysis in Section~\ref{sec:error}.

We choose Arabic as our experimental playground since it affords a rich context of linguistic variation: In addition to the standard variety, MSA, Arabic also has several dialects, thus offering an excellent context for studying our problem. From a geopolitical perspective, Arabic also has a strategic significance. This is a function of Arabic being the native tongue of $~$ 400 million speakers in 22 countries, spanning across two continents (Africa and Asia). In addition, the three dialects of our choice (EGY, GLF, LEV) are popular dialects that are widely used online. This makes our resulting models highly useful in practical situations at scale. Pragmatically, ability to develop NLP systems on dialectal tasks with no-to-small labeled dialect data immediately eases a serious bottleneck. Arabic dialects differ among themselves and from MSA at all linguistic levels, posing challenges to traditional NLP approaches. Having to develop annotated resources across the various dialects for the different tasks would be quite costly, and perhaps unnecessary. Therefore, zero-shot cross-dialectal transfer would be valuable when only some language varieties have the labeled resources.
We also note that our method is language-independent, and we hypothesize it can be directly applied to other varieties of Arabic or in other linguistic contexts for other languages and varieties.

Our research contributions in this paper are 3-fold:

\begin{enumerate}
    \item We study the problem of MSA-to-DA transfer in the context of sequence labeling and text classification and show, through experiments, that when training with MSA data only, a wide performance gap exists between testing on MSA and DA. That is, models fine-tuned on MSA generalize poorly to DA in zero-shot settings.
    
    \item We propose self-training to improve zero- and few-shot MSA-to-DA transfer. Our approach requires little-to-no labeled DA data. We evaluate extensively on 3 different dialects across the 3 aforementioned tasks, and show that our method indeed narrows the performance gap between MSA and DA by a margin as wide as $\sim$ 10\% F$_1$ points. Moreover, we conduct an ablation experiment to evaluate the importance of using unlabeled DA rather than MSA data in the zero-shot setting, and we show that unlabeled DA data is indeed much more effective and necessary for adapting the model to DA data during testing.
    
    \item We develop state-of-the-art models for the 3 tasks of (NER, POS tagging, and SRD), which we intend to publicly release for the research community.
\end{enumerate}

We now review relevant literature.

\section{Related Work}\label{sec:rel}

Classical machine learning techniques, including SVM and Conditional Random Fields (CRFs) \cite{wallach2004conditional} applied manually-extracted, hand-crafted  word- and character-level features, were previously employed for various sequence labeling tasks including NER, POS tagging, chunking. More recently, however, neural architectures, have become the \textit{defacto} approach for various tasks including sequence labeling. This usually includes an autoregressive architecture such as vanilla Recurrent Neural Networks (RNN) \cite{medsker2001recurrent} or the more sophisticated Long Short-Term Memory networks (LSTM) \cite{hochreiter1997long}. The networks processes the input text in a word-by-word fashion, and the network is trained to predict the correct label for each word. In addition, more capacity can be given to such networks by adding an additional layer that processes the input in a right-to-left fashion \cite{schuster1997bidirectional,huang2015bidirectional}.

Neural approaches usually make use of both word- and character- features. Word-level features usually consist in semantic word embeddings, which are trained on a large raw corpus in a self-supervised fashion \cite{mikolov2013distributed,pennington2014glove}. Character-level features can be extracted through an additional network such as LSTM ~\cite{lample2016neural} or CNN \cite{ma2016end}. Neural techniques has produced better or comparable results to classical approaches in addition to alleviating the need to manually hand-craft features.

In the context of Arabic NLP, the above neural techniques have also been applied to sequence tagging tasks including NER ~\cite{gridach2016character,khalifa2019character,al2020transfer,el2019arabic}, POS tagging \cite{alharbi2018part,alkhwiter2021part}, and segmentation \cite{samih2017neural}, outperforming classical rule-based approaches \cite{shaalan2014hybrid,darwish2013named}, which certainly shows the promise of these techniques when applied to morphologically-rich languages such as Arabic.

With respect to \textbf{NER} but mostly in the context of MSA, due to lack of dialectal NER datasets. For example, \cite{abdul2010simplified} applied a CRF layer over n-gram features to perform NER. \cite{abdallah2012integrating} combined a decision tree \cite{song2015decision} with rule-based features. Other, but little, work has focused on NER in the context of social media data, where DA and MSA are usually mixed together. For instance, \cite{darwish2013named} used cross-lingual resources, namely English to improve Arabic NER. However, they obtained poor results when evaluating on social media data. More recently,    \cite{gridach2016character} applied bi-directional LSTM networks on both character- and word-levels to perform NER on the Tweets dataset \cite{darwish2013named}. As for Egyptian dialect, specifically, \cite{zirikly2015named} performed NER by applying a CRF tagger on a set of lexical, morphological, and gazetteer-based features. Their approach showed improvements over baselines but the performance on dialectal data was not on par with it on MSA data, showing the challenges brought by dialectal contexts. To the best of our knowledge, little attention has been given to NER on dialectal Arabic and no prior work has studied the performance when training on MSA data and evaluating on DA data, respectively. 

As for \textbf{POS tagging} and similarly to NER, the performance of models trained on MSA drops significantly when used with DA \cite{pasha2014madamira,alharbi2018part}. Initial systems for Arabic POS tagging relied on both statistical features and linguistic rules crafted by experts \cite{khoja2001apt,alqrainy2008morphological} or combined machine learning techniques with rules \cite{tlili2006hybrid}. More recent work adopted classical machine learning model such as SVM applied on n-gram features \cite{diab2004automatic,yousif2008arabic}. Other work used n-gram features. RNNs and their variants were later adapted for the task \cite{darwish2017arabic,alharbi2018part,alrajhi2019automatic}.   

Dialectal Arabic POS tagging has received some attention although usually limited to work individual dialects such as Gulf \cite{khalifa2017morphological,alharbi2018part} and Egyptian \cite{duh2005pos,al2012yadac}. \cite{darwish2018multi} studied multi-dialectal POS tagging by proposing an annotated DA dataset from twitter spanning 4 different dialects, namely, Gulf, Egyptian, Levantine, and Maghrebi. While their results show a performance drop on DA when training on MSA only, no attempt was done to improve the DA performance in that case. We can see that despite both the difficulty and scarcity of annotated DA data for all of the different dialects and tasks, most previous work has focused on annotating uni-dialectal datasets attempting to leverage the already abundant MSA datasets. A classical work \cite{duh2005pos}, who employed an MSA morphological analyzer with a minimal supervision to perform POS tagging on Egyptian data with unlabeled Egyptian and Levantine data.

\textbf{Sarcasm Detection} (SRD) is the task of identifying sarcastic utterances where the author intends a different meaning than what is being literally enunciated \cite{van2018semeval}. Sarcasm detection is crucial for NLU as neglecting sarcasm can easily lead to the misinterpretation of the intended meaning, and therefore significantly degrade the accuracy of tasks such as sentiment classification, emotion recognition, and opinion mining. Much research effort has addressed Sarcasm detection in English, where abundant resources exist \cite{barbieri2014modelling,abercrombie2016putting,joshi2015harnessing,bouazizi2016pattern}. Earlier methods employed linguistic rules \cite{bharti2015parsing} or classical machine learning models \cite{joshi2015harnessing, saha2017proposed}. More recent methods used neural networks \cite{porwal2018sarcasm,ren2018context,mandal2019deep,jain2020sarcasm,kumar2020sarcasm,he2020sarcasm} or pre-trained language models \cite{baruah2020context,srivastava2020novel,kumar2021adversarial,potamias2020transformer}.

With respect to Arabic Sarcasm Detection, the majority of research has focused on detecting sarcastic tweets. \cite{karoui2017soukhria} used Random Forests to identify sarcastic political tweets. \cite{ghanem2019idat} proposed a shared task on irony detection in Arabic Tweets. The submitted systems to the shared task varied in their approaches from classical models with count-based features \cite{khalifa2019ensemble,nayel2019benha} to deep models \cite{ranasinghe2019rgcl,ZhangA19a}. \cite{farha2020arabic} highlighted the connection between sentiment analysis and sarcasm detection, by showing how sentiment classifiers fail with sarcastic inputs. They also proposed the largest publicly available Arabic sarcasm detection dataset, ArSarcasm, which we use in this work. We can see that so far, sarcasm detection methods have been applied to social media data collectively, with no effort made to study the zero-shot performance across dialects of state-of-the-art methods.


\textbf{Pre-trained Language Models.}
Sequential transfer learning, where a network is first pre-trained on a relevant task before fine-tuning on the target task, originally appeared in domain of computer vision, and has recently been adapted in NLP. \cite{howard2018universal} Proposed to pre-train a LSTM network for language modeling and then fine-tune for classification. Similarly, ELMO \cite{DBLP:conf/naacl/PetersNIGCLZ18} leveraged contextual representations obtained from a network pretrained for language modeling to perform many NLP tasks. Similar approaches were proposed such as BERT \cite{devlin2018bert} that relied not on RNNs, but on bidirectional Transformers \cite{vaswani2017attention}, and on a different pre-training objective, namely masked language modeling. Other variations appeared including RoBERTa~\cite{liu2019roberta}, MASS~\cite{song2019mass}, and ELECTRA~\cite{clark2020electra}. Fine-tuning these pre-trained models on task-specific data has produced state-of-the-art performance, especially in cases when sufficiently large labeled data does not exist. They have been applied to several tasks, including text classification, question answering, named entity recognition~\cite{conneau2019unsupervised}, and POS tagging~\cite{tsai2019small}. 

\textbf{Cross-lingual Learning.} Cross-lingual learning (CLL) refers to using labeled resources from resource-rich languages to build models for data-scarce languages. In a sense, knowledge learned about language structure and tasks is \textit{transferred} to low-resource languages Cross-lingual learning is of particular importance due to the scarcity of labeled resources in many of the world's languages, some of which are spoken by millions of people (Marathi and Gondi, for example). While our work can be better described as cross-dialectal, the techniques used for cross-lingual learning can easily be adapted for settings such as ours. In this work, Modern Standard Arabic (MSA) and Arabic dialects (DA) represent the high-resource and low-resource languages, respectively.

Many techniques were proposed for CLL, including using cross-lingual word embeddings \cite{ruder2019survey,adams2017cross,wang2017multi,xie2018neural}, where the two monolingual vector spaces are mapped into the same shared space. While cross-lingual word embeddings enable comparing meaning across languages \cite{ruder2019survey}, they typically fail when we do not have enough data to train good monolingual embeddings.
In addition, adversarial learning \cite{goodfellow2014} has played an important role in cross-lingual learning where an adversarial objective is employed to learn language-independent representations ~\cite{barone2016towards,kim2017cross,chen2018adversarial,DBLP:conf/emnlp/KeungLB19}. As a result, the model learns to rely more on general language structure and commonalities between languages, and therefore can generalize across languages.
Multilingual extensions of pre-trained language models have emerged through joint pre-training on several languages. Examples include mBERT~\cite{devlin2018bert}, XLM~\cite{lample2019cross} and XLM-RoBERTa~\cite{conneau2019unsupervised}. During pre-training on multiple languages, the model learns to exploit common structure among pre-training languages even without explicit alignment \cite{WuD19}. These models have become useful for few-shot and zero-shot cross-lingual settings, where there is little or no access to labeled data in the target language. For instance~\cite{conneau2019unsupervised} evaluate a cross-lingual version of RoBERTa~\cite{liu2019roberta}, namely XLM-RoBERTa, on cross-lingual learning across different tasks such as question answering, text classification, and named entity recognition. 

\textbf{Semi-supervised Learning.} Several methods were proposed for leveraging unlabeled data for learning including co-training \cite{blum1998combining}, graph-based learning \cite{culp2007graph}, tri-training \cite{zhou2005tri}, and self-training \cite{nigam2000analyzing}. A variety of semi-supervised learning methods have been successfully applied to a number of NLP tasks including  NER~\cite{kozareva2005self,helwe2019arabic}, POS tagging~\cite{wang2007semi}, parsing~\cite{sagae2010self}, word sense disambiguation \cite{mihalcea2004co}, and text classification~\cite{kiritchenko2001email,van2016predicting}. Self-training has been applied in cross-lingual settings where gold labels are rare in the target language. For example, ~\cite{hajmohammadi2015combination} proposed a combination of Active learning and self-training for cross-lingual sentiment classification.~\cite{pan2017cross} made use of self-training for named entity tagging and linking across 282 different languages. \cite{artetexe2019} used self-training for cross-lingual word mapping to create additional word pairs for training. ~\cite{dong2019robust} employed self-training to improve zero-shot cross-lingual sentiment classification with mBERT \cite{devlin2018bert}. With English as their source language, they improved performance on 7 languages by self-training using unlabeled data in their target languages. Lastly, \cite{dong2020leveraging} used the self-labeled examples produced by self-training to create adversarial examples in order to improve robustness and generalization.

We now introduce our tasks.

\section{Tasks}
Named Entity Recognition (NER) is defined as the information extraction task that attempts to locate, extract, and automatically classify named entities into predefined classes or types in unstructured texts \cite{nadeau2007survey}. Typically, NER is integrated into more complex tasks, where, for example, we might need to handle entities in a special way. For instance, when translating the Arabic sentence ``\<حقق كرم فضية المصارعه>'' to English, it would be useful to know that ``\<كرم>'' is a person name, and therefore should not be be translated into the word ``generosity''. Similarly, NER can be useful for other tasks question answering, information retrieval and summarization. 

Part-of-Speech (POS) tagging is the task of assigning a word in a context to its part-of-speech tag. Such tags include adverb (ADV), adjective (ADJ), pronoun (PRON), and many others. For example, given an input sentence ``\<أنا أحب كرة القدم>'', our goal is to tag each word as follows: \<أنا> (PRON) \<أحب> (VERB) \<كرة> (NOUN) \<ال> (DET) \<قدم> (NOUN). POS tagging is an essential NLU task with many applications in speech recognition, machine translation, and information retrieval. Both NER and POS tagging are sequence labeling tasks, where we assign a label to each word in the input context.

Sarcasm Detection is the task of identifying sarcastic utterances where the author intends a different meaning than what is being literally enunciated \cite{van2018semeval}. Sarcasm detection is crucial for NLU as neglecting to detect sarcasm can easily lead to the misinterpretation of the intended meaning, and therefore significantly degrade the accuracy of tasks such as sentiment classification, emotion recognition, and opinion mining \cite{farha2020arabic}. For example the word ``\<سعيد>'' in the utterance ``\< أنا سعيد جدا بهذا الجوال البطيء>'' can erroneously lead sentiment classifiers into positive sentiment, although the sentiment has negative sentiment.
Sarcasm Detection is typically treated as a binary classification task, where an utterance is classified as either sarcastic or not. 

\section{Method}~\label{sec:method}
In this work, we show that models trained on MSA for NER, POS tagging, and Sarcasm Detection generalize poorly to dialect inputs when used in zero-shot-settings (i.e., no annotated DA data used during training). Across the three tasks, we test how self-training would fare as an approach to leverage unlabeled DA data to improve performance on DA. Self-training involves training a model using its own predictions on a set of unlabeled data identical from its original training split. Next, we formally describe our algorithm. The notation used in this section to describe our algorithm is directed towards sequence labeling (since we experiment with 2 sequence labeling tasks out of 3). However, it should be straightforward to adapt it to the context of text classification as in \cite{dong2019robust}. 

\subsection{Self-training for Sequence Labeling}
\begin{figure}[t!] 
\centering
\includegraphics[width=6cm, height=5.1cm]{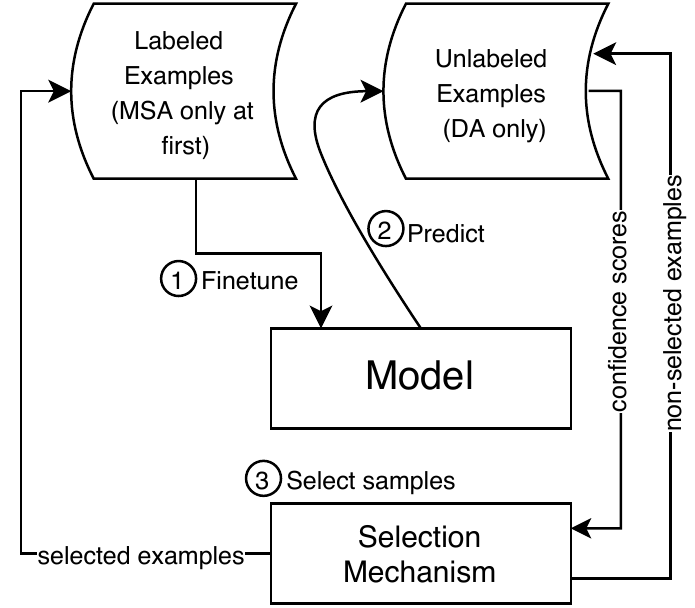}
\caption{MSA-to-DA Self-training transfer.}
\label{Fig:Model}
\end{figure}

For sequence labeling, our proposed self-training procedure is given two sets of examples: a labeled set $L$ and an unlabeled set $U$. To perform zero-shot MSA-to-DA transfer, MSA examples are used as the labeled set, while unlabeled DA examples are the unlabeled set. As shown in Figure ~\ref{Fig:Model}, each iteration of the self-training algorithm consists mainly in three steps. First, a pre-trained language model is fine-tuned on the labeled MSA examples $L$. Second, for every unlabeled DA example $u_i$, we use the model to tag each of its tokens to obtain a set of predictions and confidence scores for each token $p_{u_i} = (l^{(i)}_{1}, c^{(i)}_{1}), (l^{(i)}_{2}, c^{(i)}_{2}), ... (l^{(i)}_{|u_i|}, c^{(i)}_{|u_i|})$, where $(l^{(i)}_{j}, c^{(i)}_{j})$ are the label and confidence score (Softmax probability) for the $j$-th token in $u_i$. Third, we employ a selection mechanism to identify examples from $U$ that are going to be added to $L$ for the next iteration.

\begin{algorithm}[ht]
\SetAlgoLined
\textbf{Given} set $L$ of labeled MSA examples, set $U$ of unlabeled DA examples, $\tau$ parameter for probability threshold selection.

\textbf{repeat}

  \Indp Fine-tune model $M$ for $K$ epochs on labeled MSA examples $L$;\\
  
  \For{$u_i \in U$}{
  Obtain prediction $p_{u_i}$ on unlabeled DA example $u_i$ using model $M$;\\
  \uIf{$\min \limits_{(l^{(i)}_{j}, c^{(i)}_{j}) \in p_{u_i}}{c^{(i)}_j} \geq \tau$}{  remove $u_i$ from $U$ and add it to $L$;\\
  }
  }
  \Indm
  \textbf{until} stopping criterion satisfied
 \caption{\label{alg}MSA-to-DA Self-Training for Sequence Labeling}
\end{algorithm}

For a selection mechanism, we experiment with both a thresholding approach and a fixed-size~\cite{dong2019robust} approach. In the thresholding method, a threshold $\tau$ is applied on the minimum confidence per example. That is, we only add an example $u_i$ to $L$ if $\min \limits_{(l^{(i)}_{j}, c^{(i)}_{j}) \in p_{u_i}}{c^{(i)}_j} \geq \tau$. See Algorithm~\ref{alg}. The fixed-size approach involves, at each iteration, the selection of the top $S$ examples with respect to the minimum confidence score $\min \limits_{(l^{(i)}_{j}, c^{(i)}_{j}) \in p_{u_i}}{c^{(i)}_j}$ , where $S$ is a hyper-parameter. We experiment with both approaches and report results in Section~\ref{sec:res}. 

\subsection{Self-training for Classification}
For sarcasm detection, we follow \cite{dong2019robust} who select an equivalent number of examples from each class, which we will refer to as \textit{class balancing}. In other words, let $c_{u_i}$ be the confidence of the most probable class assigned to example $u_i$. Then we sort the unlabeled examples in a descending order according to their confidence and select the top $\floor{S/C}$ examples from each class such that we have a total of $S$ examples, where $C$ is the number of classes.

For example if $S=100$ and $C=2$ i.e we have 2 classes, we will select the top 50 confident examples that were classified as positive and the top 50 confident examples classified as negative. Similarily to \cite{dong2019robust}, we observe the positive effect of class balancing on the performance of self-training in sarcasm detection\footnote{We do not use class balancing with sequence labeling tasks since each example contains a set of tokens, each assigned to a possibly different class, which makes it very difficult to guarantee that an equal number of examples are selected for each class.} and we compare class balancing against selecting the top $S$ confident example regardless of their predicted class. See section ~\ref{sec:msa-da-zero-st}.

\begin{algorithm}[ht]
\SetAlgoLined
\textbf{Given} set $L$ of labeled MSA examples, set $U$ of unlabeled DA examples, $S$ total number of unlabeled examples to add to the training data every iteration, $C$ the number of classes.

\textbf{repeat}

  \Indp Fine-tune model $M$ for $K$ epochs on labeled MSA examples $L$;\\
  
  Obtain class predictions and confidences on all unlabeled DA examples $u_i$ using model $M$;\\
  Sort all unlabeled examples $u_i$ in descending order by the confidence of their most probable class $c_{u_i}$;\\
  Select the top $\floor{S/C}$ examples from each class, remove them from $U$, and add them to $L$;
  
  \Indm
  \textbf{until} stopping criterion satisfied
 \caption{\label{alg-clf}MSA-to-DA Self-Training for Classification}
\end{algorithm}

\section{Pretrained Language Model} 
In this work, we turn our attention to fine-tuning pre-trained language models (PLMs) on our three tasks. While self-training can basically be applied to many types of other models such as LSTM networks \cite{hochreiter1997long}, we select PLMs for two reasons. First, PLMs have been shown to outperform models trained from scratch on a wide variety of tasks \cite{devlin2018bert,howard2018universal,lample2019cross}. Second, we aim to show that even state-of-the-art models still perform poorly in certain low-resource settings asserting that we still need methods to handle such scenarios.

Pre-trained language models make use
As a pre-trained language model, we use XLM-RoBERTa~\cite{conneau2019unsupervised} (XML-R for short). XLM-R is a cross-lingual model, and we choose it since it is reported to perform better than mBERT, the multilingual model from Google~\cite{devlin2018bert}. XLM-R also uses Common Crawl for training, which is more likely to have dialectal data than Wikipedia Arabic (used in mBERT), making it more suited to our work. We now introduce our experiments.

\section{Experiments}\label{sec:exps}

We begin our experiments with evaluating the standard fine-tuning performance of XLM-R models on NER, POS tagging, and SRD against strong baselines. We then use our best models from this first round to investigate the MSA-to-DA zero-shot transfer, showing a significant performance drop even when using pre-trained XLM-R. Consequently, we evaluate self-training in zero- (NER, POS tagging, SRD) and few-shot (POS tagging) settings, showing substantial performance improvements in both cases. We now introduce our datasets.


\subsection{Datasets}~\label{subsec:data}
\noindent 
\textbf{NER:} For our work on NER, we use 4 datasets: ANERCorp~\cite{Benajiba2007};ACE 2003 ~\cite{mitchell2003tides} BNews (BN-2003);ACE 2003 Newswire (NW-2003); and Twitter~\cite{darwish2013named}. Named entity types in all datasets are \textit{location (LOC)}, \textit{organization (ORG)}, and \textit{person (PER)}.

\textbf{POS Tagging:} There are a number of Arabic POS tagging datasets, mostly on MSA~\cite{maamouri2004penn} but also on dialects such as EGY~\cite{maamouri2014developing}. To show that the proposed approach is able to work across multiple dialects, we ideally needed data from more than one dialect. Hence, we use the multi-dialectal (MD) dataset
from~\cite{darwish2018multi}, comprising 350 tweets from various Arabic dialects including MSA, Egyptian (EGY), Gulf (GLF), and Levantine (LEV). This dataset has 21 POS tags, some of which are suited to social media (since it is derived from Twitter). We show the POS tag set from ~\cite{darwish2018multi} in Table~\ref{Tab:app-pos-tags} (in the Appendix). We further evaluate fine-tuning XLMR for POS tagging on a Classical Arabic dataset, namely the Quranic Arabic Corpus (QAS). \cite{dukes2010morphological}.


\textbf{Sarcasm Detection:} We use the Ar-Sarcasm dataset provided by ~\cite{farha2020arabic}, which has a total of 10,547 example split into training and test sets. Each example in this dataset is labeled by its dialect and sarcasm label. For our experiments, we set aside 20\% of the training data as a development set. Table ~\ref{Tab:datasets} shows sizes of the datasets used. We now introduce our baselines.

\begin{table*}[h!]
\begin{center}
\begin{tabular}{lll}
\toprule
\bf Task & \bf Dataset &  \bf Size  \\ \hline
NER & ANERCorp \cite{Benajiba2007}  & $\sim$150K tokens  \\
 & ACE 2003-BNews \cite{mitchell2003tides} &  $\sim$ 15K tokens \\
 & ACE 2003-News Wire \cite{mitchell2003tides}  &  $\sim$ 27K tokens \\
  & ACE 2004-News Wire \cite{mitchell2003tides}  &  $\sim$ 70K tokens \\
 & Twitter  \cite{darwish2013named} &  $\sim$ 81K tokens \\
\bottomrule
POS Tagging & Multi-dialectal (MD) - MSA \cite{darwish2018multi} &  $\sim$ 26K tokens  \\
 & Multi-dialectal (MD) - EGY \cite{darwish2018multi} &  $\sim$ 23K tokens  \\
 & Multi-dialectal (MD) - GLF \cite{darwish2018multi} &  $\sim$ 21K tokens  \\
 & Multi-dialectal (MD) - LEV \cite{darwish2018multi} &  $\sim$ 23K tokens \\

& Quranic Arabic Corpus (QAC) &  $\sim$ 134K tokens  \\
\bottomrule
Sarcasm Detection & Ar-Sarcasm \cite{farha2020arabic} & $\sim$ 10K sentences  \\

\bottomrule
\end{tabular}
\end{center}
\caption{\label{Tab:datasets} Datasets used for each of the 3 tasks studied.}
\end{table*}


\subsection{Baselines}
For the \textbf{NER task}, we use the following baselines:





\begin{itemize}
\small 
    \item \textbf{NERA}\textbf{~\cite{abdallah2012integrating}}: A hybrid system of rule-based features and a decision tree classifier.
    \item \textbf{WC-BiLSTM}\textbf{~\cite{gridach2016character}}: A character- and a word-level Bi-LSTM with a conditional random fields (CRF) layer. 
    \item \textbf{WC-CNN}\textbf{~\cite{khalifa2019character}}: A character- and a word-level CNN with  a CRF layer.
    \item \textbf{mBERT}\textbf{~\cite{devlin2018bert}}: A fine-tuned multilingual BERT-Base-Cased (110M parameters), pre-trained with a masked language modeling objective on the Wikipedia corpus of 104 languages (including Arabic). For fine-tuning, we find that (based on experiments on our development set) a learning rate of $6 \times 10^{-5}$ works best with a dropout of 0.1. 
\end{itemize}
In addition, we compare to the published results in~\cite{shaalan2014hybrid}, AraBERT~\cite{antoun2020arabert}, and CAMel~\cite{obeid2020camel} for the ANERCorp dataset. We also compare to the published results in~\cite{khalifa2019character} for the 4 datasets.

For the \textbf{POS tagging task}, we compare to our own implementation of WC-BiLSTM (since there is no published research that uses this method on the task, as far as we know) and run mBERT on our data. We also compare to the CRF results published by~\cite{darwish2018multi}. In addition, for the Gulf dialect, we compare to the BiLSTM with compositional character representation and word representations (CC2W+W) published results in~\cite{alharbi2018part}.

For the \textbf{Sarcasm Detection task}:
\begin{itemize}
\small 
    \item \textbf{Word-level BiLSTM}: A bidirectional LSTM on the word level. We use the same hyper-parameters as in \cite{farha2020arabic}.
    \item \textbf{Word-level CNN \cite{DBLP:conf/emnlp/Kim14}}: the network is has one convolutional layer of 10 filters of sizes 3, 5, and 7.
    \item \textbf{mBERT}\textbf{~\cite{devlin2018bert}}: mBERT fine-tuned for SRD. Here, we find that a different learning rate of $5 \times 10^{-6}$ performs best.
\end{itemize}

\subsection{Experimental Setup}

Our main models are XLM-R\textsubscript{BASE} $(L=12, H=768, A=12, \textnormal{ 270M params})$ and XLM-R\textsubscript{LARGE} $(L=24, H=1024, A=16,\textnormal{ 550M params})$, where $L$ is number of layers, $H$ is the hidden size, $A$ is the number of self-attention heads. For XLM-R experiments, we use Adam optimizer with $1e^{-5}$ learning rate, batch size of 16. We typically fine-tune for 20 epochs, keeping the best model on the development set for testing. We report results on the test split for each dataset, across the two tasks. For all BiLSTM experiments, we use the same hyper-parameters as~\cite{khalifa2019character}.


For all the self-training experiments, we use the dialect subset of the Arabic online news commentary (AOC) dataset~\cite{zaidan2011arabic}, comprising the EGY, GLF, and LEV varieties limiting to equal sizes of 9K examples per dialect (total =27K)~\footnote{We note that our approach could be scaled with an even bigger unlabeled dataset, given the performance gains we report with self-training in this work.}. We use the split from~
\cite{elaraby2018deep} of AOC, removing the dialect labels and just using the comments themselves for our self-training. Each iteration involved fine-tuning the model for $K=5$ epochs. As a stopping criterion, we use early stopping with patience of 10 epochs. Other hyper-parameters are set as listed before. For selecting confident samples,  we experiment with a fixed number of top samples $S=[50, 100, 200]$ and selection based on a probability threshold $\tau=[0.80, 0.90, 0.95]$ (softmax values)~\footnote{It is worth noting that our $S$ values are similar to those used in~\cite{dong2019robust}. We also experimented with other values for $\tau$ and $S$, but found them sub-optimal and hence we report performance only for the listed values of these two hyper-parameters here.}. For all evaluations, we use the \textit{seqeval} toolkit~\footnote{https://github.com/chakki-works/seqeval}. 


\section{Results}\label{sec:res}
\subsection{Fine-tuning XLM-R}\label{sec:finetune}
We start by showing the result of fine-tuning XLM-R on the \textbf{NER task}, on each of the 4 Arabic NER (ANER) datasets listed in Section~\ref{subsec:data}. Table~\ref{Tab:ner} shows the test set macro F$_1$ score on each of the 4 ANER datasets.  Clearly, the fine-tuned XLM-R models outperform other baselines on all datasets, except on the NW-2003 where WC-CNN~\cite{khalifa2019character} performs slightly better than XLM-R\textsubscript{LARGE}.

For \textbf{POS Tagging}, Table ~\ref{Tab:pos} shows test set word accuracy  of the XLM-R models compared to baselines on the Quranic Arabic Corpus (QAC) and 4 different subsets from the multi-dialectal dataset \cite{darwish2018multi}. Again, XLM-R models (both base and large) outperform all other models. A question arises why XLM-R models outperform both mBERT and AraBERT. As noted before, for XLM-R vs. mBERT, XLM-R was pre-trained on much larger data: CommonCrawl for XLM-R vs. Wikipedia for mBERT. Hence, the \textit{larger dataset} of XLM-R is giving it an advantage over mBERT. For comparison with AraBERT, although the pre-training data for XLM-R and AraBERT may be comparable, even the smaller XLM-R model (XLM-R\textsubscript{BASE}) has more than twice the number of parameters of the BERT\textsubscript{BASE} architecture on which AraBERT and mBERT are built (270M v. 110M). Hence, XLM-R model \textit{capacity} gives it another advantage. We now report our experiments with zero-shot transfer from MSA to DA.

For \textbf{Sarcasm Detection}, we fine-tune XLM-R\textsubscript{BASE} and XLM-R\textsubscript{LARGE} on the full Ar-Sarcasm dataset and compare their performance against three different baselines in Table ~\ref{Tab:sd-finetune}. Worst performance is given by CNN, which can be attributed to the way CNNs work; by capturing local n-gram features, the CNN filters fail to learn the wide contextual features required to detect sarcasm. Clearly, mBERT is performing very well compared to BiLSTM and CNN but XLM-R\textsubscript{BASE} and XLM-R\textsubscript{LARGE} outperfrom all other baselines on the task with 69.83\% and 74.07\% macro F1 points, respectively, achieving new state-of-the-art on the Ar-Sarcasm dataset. 

\begin{table*}[h]
\begin{center}
\begin{tabular}{llllll}
\toprule
\bf Model &  \bf ANERCorp &  \bf BN-2003 & \bf NW-2003 & \bf NW-2004 &  \bf Twitter \\ \hline
NERA \cite{abdallah2012integrating} & 88.77 & ~ ~ --  & ~ ~ -- & ~ ~ --  &  ~ ~ --    \\
CAMeL \cite{obeid2020camel} & 85.00 & ~ ~ --  & ~ ~ --  &  ~ ~ --  &  ~ ~ --     \\
Hybrid \cite{shaalan2014hybrid} & 90.66 & ~ ~ --  & ~ ~ --  &  ~ ~ -- &  ~ ~ --      \\
WC-BiLSTM \cite{gridach2016character}  & 88.56  & 94.92 & 90.32 & 89.62 & 64.93   \\
WC-CNN \cite{khalifa2019character} & 88.77 & 94.12 & \bf 91.20 & \bf 91.47 & 65.34   \\
mBERT (ours) & 85.86 & 89.52  &  87.19  & 88.58 & 58.92 \\
AraBERT \cite{antoun2020arabert} & 84.2 & ~ ~ --  & ~ ~ --  &  ~ ~ --   &  ~ ~ --  \\

\hline
XLM-R\textsubscript{BASE} (ours) & 87.75 & 95.35 & 85.25 & 89.61 & 60.39   \\
XLM-R\textsubscript{LARGE} (ours) &  \textbf{91.43} & \textbf{97.33} & 91.10 & 90.78 &  \textbf{68.91}   \\

\bottomrule
\end{tabular}
\end{center}
\caption{\label{Tab:ner} Test set macro F$_1$ scores for NER.}
\end{table*}

\begin{table*}[h]
\begin{center}
\begin{tabular}{llllll}
\toprule
\bf Model &  \bf QAC &  \bf MD-MSA & \bf MD-EGY & \bf MD-GLF & \bf MD-LEV  \\ \hline
BiLSTM (CC2W + W) \cite{alharbi2018part} & ~~-- & ~~--  &  ~~--  & 89.7 & ~~--  \\
CRF \cite{darwish2018multi} & ~ ~ --   &  93.6  & 92.9 & 87.8 & 87.9 \\
WC-BiLSTM  (ours) & 91.65   & 94.63  & 93.41 & 88.79 & 86.13\\
mBERT (ours) & 94.83 & 90.57  &  92.88  & 87.85 & 72.30  \\
\hline 
XLM-R\textsubscript{BASE} (ours)  & \textbf{96.70} & 96.30 & 94.70 & 92.18 & 89.98\\
XLM-R\textsubscript{LARGE} (ours) & 96.59  & \bf 98.21 & \bf 97.00 & \bf 94.41 & \bf 93.19\\
\bottomrule
\end{tabular}
\end{center}
\caption{\label{Tab:pos} Test set accuracy for POS Tagging using several baselines.}
\end{table*}

\begin{table}[h]
\begin{center}
\begin{tabular}{lll}
\toprule
\bf Model &  \bf  DEV & \bf TEST  \\ \hline
BiLSTM \cite{farha2020arabic} & 63.51  &  62.19  \\
CNN  & 59.7  & 58.50  \\
mBERT & 68.87  & 69.51 \\
XLM-R\textsubscript{BASE} (ours)  & 73.22 & 69.83 \\
XLM-R\textsubscript{LARGE} (ours)  & \bf 73.72 & \bf 74.07 \\
\bottomrule
\end{tabular}
\end{center}
\caption{\label{Tab:sd-finetune} Macro F1 scores with several baselines for sarcasm detection. Dataset used is Ar-Sarcasm. \cite{farha2020arabic}}

\end{table}

\subsection{MSA-DA Zero-Shot Transfer}\label{sec:msa-da-zero}
As before, we start by the discussion of \textbf{NER experiments}. To evaluate the utility of approach, we obviously need DA data labeled for NER. We observed that the dataset from ~\cite{darwish2013named} contains both MSA and DA examples (tweets). Hence, we train a binary classifier 
to distinguish DA data from MSA\footnote{The classifier is XLM-R\textsubscript{BASE} fine-tuned on the AOC data. The fine-tuned model achieved development and test accuracies of 90.3\% and 89.4 \%, respectively, outperforming the best results in \cite{elaraby2018deep}.}. We then extract examples that are labeled with probability $p > 0.90$ as either DA or MSA. We obtain 2,027 MSA examples (henceforth, \textit{Darwish-MSA}) and 1,695 DA examples (henceforth, \textit{Darwish-DA}), respectively. We split these into development and test sets with 30\% and 70\% ratios. As \textbf{for POS Tagging}, we already have MSA data for training and the three previously used DA datasets, namely EGY, GLF and LEV, for evaluation. We use those for the zero-shot setting by omitting their training sets and using only the development and test sets.


We first study how well models trained for NER and POS tagging on MSA data only will generalize to DA inputs during test time. We evaluate this zero-shot performance on both the XLM-R\textsubscript{BASE} and XLM-R\textsubscript{LARGE} models. \textbf{For NER}, we train on ANERCorp (which is pure MSA) and evaluate on both Darwish-MSA and Darwish-DA. While for POS tagging, we train on the MSA subset~\cite{darwish2018multi} and evaluate on the corresponding test set for each dialect. As shown in Table ~\ref{Tab:zero-all}, For NER, a significant generalization gap of around 20 \% F$_1$ points exists between evaluation on MSA and DA using both models. While for \textbf{POS tagging}, the gap is as large as 18.13 \% accuracy for the LEV dialect with XLM-R\textsubscript{BASE}. The smallest generalization gap is on the GLF variety, which is perhaps due to the high overlap between GLF and MSA~\cite{alharbi2018part}. 

\textbf{For Sarcasm Detection}, Since Ar-Sarcasm is labeled by dialect, it is trivial to extract the MSA examples for training. Similarly to what was done with the NER data, we split all\footnote{Without this, we had only 528 and 698 development and test examples, respectively and it resulted in high variance in the results obtained. So we had to increase the sizes of the development and test sets by sacrificing the DA training data.} the remaining DA examples into development and test sets with 30\% and 70\% ratios, respectively for evaluation. Finally, we obtain 4506 MSA training, 1202 DA development, and 2268 DA test examples. As shown in Table ~\ref{Tab:zero-all}, a performance gap of around 8 macro F1 points with both XLM-R\textsubscript{BASE} and XLM-R\textsubscript{LARGE}, showing poor generalization on DA in context of text classification, as well. In the next section, we evaluate the ability of self-training to close this MSA-DA performance gap.

\begin{table*}[h]
\begin{center}
\begin{tabular}{c|c|c|c|c|c|c|c|c}
\toprule

\textbf{Model}
    & \multicolumn{2}{c|}{\textbf{NER}}
        &  \multicolumn{4}{c|}{\textbf{POS Tagging}}
        & \multicolumn{2}{c}{\textbf{Sarcasm Detection}} \\

\hline
 & \textbf{MSA} & \textbf{DA} & \textbf{MSA} &  \textbf{EGY} & \textbf{GLF} & \textbf{LEV} 
 & \textbf{MSA} & \textbf{DA} \\
\hline

XLM-R\textsubscript{BASE} & 60.42 & 40.07 & 96.30 & 78.38 & 83.72 & 78.17 & 68.68 & 60.17 \\
XLM-R\textsubscript{LARGE} & 68.32 & 47.35 & 98.21 & 82.28 & 85.95 & 81.24  & 71.55 & 62.90 \\
\bottomrule
\end{tabular}
\end{center}
\caption{\label{Tab:zero-all} Zero-shot transfer results on the DA test sets. Metrics used are macro F$_1$ for NER and Sarcasm detection, and accuracy for POS Tagging. Models are trained on MSA only and evaluated on DA. Datasets used are: Darwish-MSA and Darwish-DA \cite{darwish2013named} (NER), Multi-dialectal \cite{darwish2018multi} (POS tagging), and Ar-Sarcasm \cite{farha2020arabic} (sarcasm detection). As shown, a significant performance drop exists when training on MSA and evaluating on DA.}
\end{table*}






\subsection{Zero-shot Self-Training}\label{sec:msa-da-zero-st}
Here, \textbf{for NER}, similar to Section~\ref{sec:msa-da-zero}, we train on ANERCorp (pure MSA) and evaluate on Darwish-MSA and Darwish-DA. Table ~\ref{Tab:st-ner} shows self-training NER results employing the selection mechanisms listed in Section~\ref{sec:method}, and with different values for $S$ and $\tau$. The best improvement is achieved with the thresholding selection mechanism with a $\tau=0.90$, where we have an F$_1$ gain of 10.03 points. More generally, self-training improves zero-shot performance in all cases albeit with different F$_1$ gains. Interestingly, we find that self-training also improves test performance on MSA with the base XLM-R model. This is likely attributed to the existence of MSA content in the unlabeled AOC data. It is noteworthy, however, that the much higher-capacity large model deteriorates on MSA if self-trained (dropping from 68.32\% to 67.21\%). This shows the ability of the large model to learn representations very specific to DA when self-trained. It is also interesting to see that the best self-trained base model achieving 50.10\% F$_1$, outperforming the large model before the latter is self-trained (47.35\% in the zero-shot setting). This shows that a base self-trained model, suitable for running on terminal machines with less computational capacity, can (and in our case does) improve over a large (not-self-trained) model that needs significant computation. The fact that, when self-trained, the large model improves 15.35\% points over the base model in the zero-shot setting (55.42 vs. 40.07) is remarkable.


\begin{table}[h]
\begin{center}
\begin{tabular}{l|l|l}
\toprule
\bf Model & \bf Darw-MSA& \bf Darw-DA \\ \hline
XLM-R\textsubscript{BASE} & 61.88 & 40.07 \\ 

XLM-R\textsubscript{BASE}, ST, S=50 & 60.98  &  43.88 \\
XLM-R\textsubscript{BASE}, ST, S=100 & 61.13 &  42.01 \\
XLM-R\textsubscript{BASE}, ST, S=200 & 61.46 &  43.49 \\

XLM-R\textsubscript{BASE}, ST, $\tau=0.80$ & \textbf{63.36} &  46.97 \\
XLM-R\textsubscript{BASE}, ST, $\tau=0.90$ & 61.02 & \bf 50.10 \\
XLM-R\textsubscript{BASE}, ST, $\tau=0.95$ & 62.25 & 47.91 \\
\hline
XLM-R\textsubscript{LARGE} & \textbf{68.32} & 47.35 \\
XLM-R\textsubscript{LARGE} + ST, $\tau=0.90$ & 67.21 & \textbf{55.42} \\

\bottomrule
\end{tabular}
\end{center}
\caption{\label{Tab:st-ner} Test set macro F$_1$ in the zero-short setting for NER. Training was done on MSA data only. \textbf{ST} stands for self-training. Models were trained on  ANERCorp (pure MSA) and evaluated on Darwish-MSA and Darwish-DA extracted from the Twitter dataset ~\cite{darwish2018multi}. Self-training boosts the performance on DA data by 10\% macro F1 points with XLM-R\textsubscript{BASE} and $\tau=0.90$.}
\end{table}

As \textbf{for POS tagging}, we similarly observe consistent improvements in zero-shot transfer with self-training (Table~\ref{Tab:zero}). The best model achieves accuracy gains of 2.41\% (EGY), 1.41\% (GLF), and 1.74\% (LEV). Again, this demonstrates the utility of self-training pre-trained language models on the POS tagging task even in absence of labeled dialectal POS data (zero-shot).

For \textbf{Sarcasm Detection}, we follow \cite{dong2019robust} in balancing the examples selected in each self-training iteration through selecting an equal number of examples from each class (sarcastic and non-sarcastic). Without the balancing step, we find that the selected examples come from the most frequent class (non-sarcastic), which hurts performance since the model is learning only one class. The results for sarcasm detection are shown in Table ~\ref{Tab:sarcasm-st}, where we see that self-training adds 3\% and 2.5\% (for XLM-R\textsubscript{BASE}) and 5.9\% and 4.5\% (for XLM-R\textsubscript{LARGE}) macro F1 points on the development and test sets, respectively using the best settings for self-training ($S=100$ with class balancing). We also find that selection based on probability thresholds performs much worse than fixed-size selection, hence we omit these results. 

\begin{table}[h]
\begin{center}
\begin{tabular}{l|l|l|l|l}
\toprule
\bf Model & \bf           MSA & \bf EGY & \bf GLF & \bf LEV \\ \hline
XLM-R\textsubscript{BASE} & 96.30 & 78.38 & 83.72 & 78.17 \\ 
XLM-R\textsubscript{BASE}, ST, S=50 & ~~~ -- & \textbf{80.79} & \textbf{85.13} & \textbf{79.91}  \\
XLM-R\textsubscript{BASE}, ST, S=100 & ~~~ -- & 80.43 & 84.74 & 79.16  \\
XLM-R\textsubscript{BASE}, ST, S=200 & ~~~ -- & 78.75 & 84.21 & 79.40  \\ 
XLM-R\textsubscript{BASE}, ST, $\tau$ = 0.90 & ~~~ -- & 79.52 & 83.97 &  79.21 \\ 
XLM-R\textsubscript{BASE}, ST, $\tau$ = 0.85& ~~~ -- & 78.97 & 83.53 &  79.06   \\ 
XLM-R\textsubscript{BASE}, ST, $\tau$ = 0.80 & ~~~ -- & 78.88 & 83.72 &  78.50 \\ 
\hline
XLM-R\textsubscript{LARGE} & 98.21 & 82.28 & 85.95 & 81.24 \\
XLM-R\textsubscript{LARGE} + ST,  S=50 &  ~~~ -- & \textbf{82.65} & \textbf{87.76} & \textbf{83.70} \\


\bottomrule
\end{tabular}
\end{center}
\caption{\label{Tab:zero} Test set accuracy in the zero-shot setting for POS tagging. \textbf{ST} stands for self-training. Models were trained on the MSA data of the when training on MSA only. Self-training boosts performance of XLMR\textsubscript{BASE} by around 2\% accuracy points on different dialects with the best setting of $S=50$. }
\end{table}

\begin{table*}[h]
\begin{center}
\begin{tabular}{c|c|c|c|c}
\toprule

\textbf{Model}
   & \multicolumn{2}{c|}{\textbf{MSA}}
        &  \multicolumn{2}{c}{\textbf{DA}} 
        \\

\hline
 & \textbf{DEV} & \textbf{TEST} & \textbf{DEV} &  \textbf{TEST}\\
 \hline

XLM-R\textsubscript{BASE} & 65.64  & 68.68 &  61.66 & 60.17  \\
XLM-R\textsubscript{BASE} + ST, S=50, & ~~~ --  & ~~~ -- &  62.53 & 60.82  \\
XLM-R\textsubscript{BASE} + ST, S=100 & ~~~ --  & ~~~ -- &  61.15 & 59.46  \\
XLM-R\textsubscript{BASE} + ST, S=200  & ~~~ --  & ~~~ -- &  62.57 & 60.25  \\

XLM-R\textsubscript{BASE} + ST, S=50, class balancing & ~~~ --  & ~~~ -- &  62.49 & 59.34  \\
XLM-R\textsubscript{BASE} + ST, S=100, class balancing & ~~~ --  & ~~~ -- &  \textbf{64.72} & \textbf{62.66}  \\
XLM-R\textsubscript{BASE} + ST, S=200, class balancing & ~~~ --  & ~~~ -- &  62.89 & 59.46  \\

\bottomrule
XLM-R\textsubscript{LARGE} & 67.81  & 71.55 &  62.28 & 62.90  \\


XLM-R\textsubscript{LARGE} + ST, S=100, class balancing & ~~~ --  & ~~~ -- &  \textbf{68.21} & \textbf{67.43}  \\

\bottomrule
\end{tabular}
\end{center}
\caption{\label{Tab:sarcasm-st} Macro F$_1$ in the zero-shot setting for sarcasm detection on the Ar-Sarcasm \cite{farha2020arabic} dataset. Training was done on MSA data only. \textbf{ST:} stands for  self-training. An obvious performance boost occurs when using self-training in the best setting with $S=100$ and class balancing.}
\end{table*}

\subsection{Ablation Experiment}
Here, we conduct an ablation experiment with the NER task in order to verify our hypothesis that the performance boost primarily comes from using unlabeled DA data for self-training. By using a MSA dataset with the same size as our unlabeled DA one\footnote{We use a set of MSA tweets from the AOC dataset mentioned before.}, we can compare the performance of the self-trained model in both settings: MSA and DA unlabeled data. We run 3 different self-training experiments using 3 different values for $\tau$ using each type of unlabeled data. Results are shown in table ~\ref{Tab:st-msa}. While we find slight performance boost due to self-training even with MSA unlabeled data, the average F1 score with unlabeled DA is better by 2.67 points, showing that using unlabeled DA data for self-training has helped the model adapt to DA data during testing.
\begin{table}[h]
\begin{center}
\begin{tabular}{l|l|l}
\toprule
\bf Setting & \bf Unlbl. MSA& \bf Unlbl. DA \\ \hline
XLM-R\textsubscript{BASE}, ST, $\tau=0.80$ & 43.88 &  44.46 \\
XLM-R\textsubscript{BASE}, ST, $\tau=0.90$ & 44.69 & 47.83 \\
XLM-R\textsubscript{BASE}, ST, $\tau=0.95$ & 43.43 & 46.87 \\
\hline
Avg & 43.67 & \textbf{46.34} \\ 
\bottomrule
\end{tabular}
\end{center}
\caption{\label{Tab:st-msa} Ablation experiment with MSA unlabeled data for zero-shot NER. Development set macro F1 is shown when using both unlabeled MSA and DA data with the same size. Average performance with DA unlabeled data is higher showing the effect of unlabeled DA on the model final performance.}
\end{table}

\section{Error analysis}\label{sec:error}



\subsection{NER}
To understand why self-training the pre-trained language model improves over mere fine-tuning, we perform an error analysis. For the error analysis, we focus on the NER task where we observe a huge self-training gain. We use the development set of Darwish-DA (See section ~\ref{sec:msa-da-zero-st}) for the error analysis. We compare predictions of the standard fine-tuned XLM-R\textsubscript{BASE} model (FT) and the best performing self-training ($\tau=0.9$) model (ST) on the data. The error analysis leads to an interesting discovery: The greatest benefit from the ST model comes mostly from reducing \textit{false positives} (see Table~\ref{tab:st-fp}). In other words, self-training helps regularize the model predictions such that tokens misclassified by the original FT model as a named entities are now correctly tagged as \textit{unnamed entity} ``O". 

To understand why the ST model improves false positive rate, we manually inspect the cases it correctly identifies that were misclassified by the FT model. We show examples of these cases in Table~\ref{tab:app-fp_analysis} (in the Appendix). As the table shows, the ST model is able to identify dialectal tokens whose equivalent MSA forms can act as trigger words (usually followed by a PER named entity). We refer to this category as \textbf{\textit{false trigger words}}. An example is the word  \<نبي > ``prophet" (row 1 in Table~\ref{tab:app-fp_analysis}). A similar example that falls within this category is in row (2), where the model is confused by the  token  \<الى> ( ``who" in EGY, but ``to" in MSA and hence the wrong prediction as LOC). A second category of errors is caused by \textbf{\textit{non-standard social media language}}, such as use of letter repetitions in interjections (e.g., in row (3) in  Table~\ref{tab:app-fp_analysis}). In these cases, the FT model also assigns the class PER, but the ST model correctly identifies the tag as ``O". A third class of errors arises as a result of \textit{\textbf{out-of-MSA}} vocabulary. For example, the words in rows (4-6) are all out-of-MSA where the FT model, not knowing these, assigns the most frequent named entity label in train (PER). A fourth category of errors occurs as a result of a token that is usually part of a named entity in MSA, that otherwise functions as part of an \textit{\textbf{idiomatic expression}} in DA. Row (7) in Table~\ref{tab:app-fp_analysis} illustrates this case.


We also investigate errors shared by both the FT and ST models (errors which the ST model also could not fix). Some of these errors result from the fact that often times both MSA and DA use the same word for both person and location names. Row (1) in Table ~\ref{tab:app-shared-errors} (in the Appendix) is an example where the word ``Mubarak", name of the ex-Egypt President, is used as LOC. Other errors include \textit{out-of-MSA} tokens mistaken as named entities. An example is in row (3) in Table ~\ref{tab:app-shared-errors}, where \<بأمارة> ,(``proof" or  ``basis" in EGY) is confused for \<بإمارة> (``emirate", which is a location). \textit{False trigger words}, mentioned before, also play a role here. An example is in row (7) where \<يابطل> is confused for  PER due to the trigger word \<يا> ``Hey!" that is usually followed by a person name. \textit{\textbf{Spelling mistakes}} cause third source of errors, as in row (4). We also note that even with self-training, detecting ORG entities is more challenging than PER or LOC. The problem becomes harder when such organizations are not seen in training such as in rows (8) \<الاخوان المسلمين>, (9) \<قناة العربية> and  (10) \<المجلس العسكري>, all of which do not occur in the training set (ANERCorp). 

Here we investigate the false negatives produces by the self-trained models observing a number of named entities that were misclassified by the self-trained model as unnamed ones. See Table ~\ref{tab:app-fn-analysis} (in the Appendix). As an example, we take the last name \<الجنزوري> which was classified both correctly and incorrectly in different contexts by the self-trained model. Context of correct classification is ``\<هاش تاج لكمال الجنزوري> ", while it is ``\<ماسك على الناس كلها سي دي الا الجنزوري ماسك عليه فلوبي>" for the incorrect classification. First, we note that \<الجنزوري> is not a common name (zero occurrences in the MSA training set). Second, we observe that in the correct case, the word was preceded by the first name \<كمال> which was correctly classified as PER, making it easier for the model to assign PER to the word afterwards as a surname.

\begin{table}[ht!]
    \centering

    \begin{tabular}{c|c|c|c}
        \toprule
        \textbf{Measure}  &  \textbf{FT} & \textbf{ST} & \textbf{\% improvement} \\
        \hline
        True Positives & 155 & 165 & +6.5 \% \\
        False Positive & 159 & 64 & +59.7 \%\\
        False Negatives & 162 & 168 & -3.7 \% \\
        True Negatives & 5,940 & 6,035 & +1.5 \% \\
        \bottomrule
    \end{tabular}
    \caption{Comparison of error categories in percentage between the fine-tuned model (FT) and the model combining fine-tuned+self-trained (ST) model for NER. The values are based on the dialectal part of the development set. }
    \label{tab:st-fp}
\end{table}



\begin{table}[h!]
    \footnotesize 
    \centering

     \begin{tabular}{l|c|c|c|c}
        \toprule
        \textbf{no.} & \textbf{Word}  & \textbf{Gold} & \textbf{FT} & \textbf{ST}  \\
        \hline
        (1) & \<الاخوان > &  ORG  &  ORG  & O \\
        (2) &   \<للبرادعي > &   PER &   PER &  O \\
         (3) &  \< مجدي الجلاد > &  PER &   PER  & O \\
        (4) &   \<فان ديزل>  &  PER &   PER  &  O \\
        (5) &  \<الجنزوري> &  PER &   PER  &  O \\
        (6) &  \<زين يسون>   &  PER &   PER  &  O \\
         \bottomrule
    \end{tabular}
    
    \caption{\textbf{NER task.} Sample false negatives produced by self-training.}
    \label{tab:fn-analysis}
\end{table}

\subsection{Sarcasm Detection}
We also conduct an error analysis on Sarcasm Detection comparing the predictions of XLM-R\textsubscript{BASE} with and without self-training. For that we use the best model on the development set (XLM-R\textsubscript{BASE}, S=100 with class balancing). Our analysis with SRD yields a similar observation to NER, where the performance boost driven by self-training is mostly due to the alleviation of false positives or the improvement of true negatives\footnote{We can see that in binary classification, every false positive removed is a true negative added}. Table~\ref{tab:st-fp-sd} compares performance measures between the two models. However, we can see that, unlike NER, false negatives increase by as much as 44\%, which is likely due to the self-training regularization effect mentioned earlier.

We also analyze sample errors that were fixed by the self-trained model. See Table ~\ref{tab:app-SRD-fixed-errors} (in the Appendix). The first four examples represent false negatives, where the fine-tuned model assumed to be non-sarcastic. We can see that in such dialectal contexts, the fine-tuned model suffers from many unseen words during training on MSA. More specifically, words such as \<بيه> and \<غساله> in example (1), or \<عاهات> in (2), \<عبيط> in (4), or an idiom such as \<حاميها حراميها> n e(3), or \<ما كانش حد غلب> in (5), or \<ياترى> in (6), all of which represent dialect-specific language that is not encountered in MSA contexts, and therefore represents a significant challenge in zero-shot settings.

In addition, we show sample errors shared between the fine-tuned and the self-training models. See Table ~\ref{tab:app-SRD-shared-errors} (in the Appendix). As to why the self-trained model has not corrected these errors, we can hypothesize that it may be due to that the vocabulary used in these inputs was not seen during self-training. In other words, this vocabulary was either not selected by the self-training selection mechanism to be added to the training data or not existing at all in the unlabeled examples used for self-training. As a result, the model was not adapted sufficiently to handle these or similar contexts. We assume the performance on these inputs could improve with larger and more diverse unlabeled examples used for self-training.

\begin{table}[ht!]
    \centering

    \begin{tabular}{c|c|c|c}
        \toprule
        \textbf{Measure}  &  \textbf{FT} & \textbf{ST} & \textbf{\% improvement} \\
        \hline
        True Positives & 737 & 688 & -6.6 \% \\
        False Positive & 230 & 185 & +19.7 \%\\
        False Negatives & 111 & 160 & -44.1\% \\
        True Negatives & 124 & 169 & +36.29 \% \\
        \bottomrule
    \end{tabular}
    \caption{Comparison of error categories in percentage between the fine-tuned model (FT) and the model combining fine-tuned+self-trained (ST) model for Sarcasm Detection, based on the dialectal part of the development set.}
    \label{tab:st-fp-sd}
\end{table}

\section{Conclusion}\label{sec:conc}
Even though pre-trained language models have improved many NLP tasks, they still need a significant amount of labeled data for high-performance fine-tuning. In this paper, we proposed to self-train pre-trained language models by using unlabeled Dialectal Arabic (DA) data to improve zero-shot performance when training on Modern Standard Arabic (MSA) data only. Our experiments showed substantial performance gains on two sequence labeling tasks (NER and POS), and one text classification task (sarcasm detection) on different Arabic varieties. Our method is dialect- and task-agnostic, and we believe it can be applied to other tasks and dialectal varieties. We intend to test this claim in future research. Moreover, we evaluated the fine-tuning of the recent XLM-RoBERTa language models, establishing new state-of-the-art results on all of the three tasks studied. 





\bibliographystyle{IEEEtran}
\bibliography{ref}

\section{Appendix}
\input{appendix}

%



\end{document}

%% file: appendix.tex
\section{POS Tag Set}\label{sec:pos-tagset}

\begin{table}[h!]
\begin{center}
\begin{tabular}{llll}
\toprule
\bf Tag &  \bf Description &  \bf Tag & \bf Description  \\ 
\hline
ADV & adverb & ADJ & adjective    \\
CONJ & conjunction & DET & determiner  \\
NOUN  & noun & NSUFF & noun suffix    \\
NUM   & number  & PART & particle  \\
PUNC & punctuation  & PRON & pronoun \\
PREP & preposition & V & verb   \\
ABBREV & abbreviation &  VSUFF  & verb suffix \\
FOREIGN & non-Arabic & FUT\_PART & future particle   \\
PROG\_PART & progressive particle & EMOT & Emoticon/Emoji\\
MENTION & twitter mention & HASH & Hashtag \\
URL & URL & ~~~ -- & ~~~ --\\
\bottomrule
\end{tabular}
\end{center}
\caption{\label{Tab:app-pos-tags} The POS tag set in~\cite{darwish2018multi}.}
\end{table}

\section{Error Analysis}
The ``regularizing" effect caused by self-training and discussed in section ~\ref{sec:error} can sometimes produce false negatives as shown in Table ~\ref{tab:fn-analysis}. We see a number of named entities that were misclassified by the self-trained model as unnamed ones. As an example, we take the last name \<الجنزوري> which was classified both correctly and incorrectly in different contexts by the self-trained model. Context of correct classification is ``\<هاش تاج لكمال الجنزوري> ", while it is ``\<ماسك على الناس كلها سي دي الا الجنزوري ماسك عليه فلوبي>" for the incorrect classification. First, we note that \<الجنزوري> is not a common name (zero occurrences in the MSA training set). Second, we observe that in the correct case, the word was preceded by the first name \<كمال> which was correctly classified as PER, making it easier for the model to assign PER to the word afterwards as a surname.

\begin{table*}[h!]
    \centering
    \footnotesize 

    \begin{tabular}{l|c|l|c|l|c}
        \toprule
        \textbf{no.} & \textbf{Token} &\textbf{Eng.} &\textbf{MSA} & \textbf{Context/Explanation} & \textbf{FT Pred.} \\
        \hline
        (1) & \<نبي > & we want  & \<نريد>  &  \< نبي نعرف من...>    (\textit{we want to know who}) & PER  \\

       (2) &   \<ماكانوا> & wasn't &  \<لم يكونوا>
          & \<أغلب الي ماكانوا مصدقين>
          (\textit{most of those who wasn't believing}) & LOC  \\

         (3) & \<لوووول > & LOL &  \<ضحك>  & \< لوووول...> (interjection) & PER  \\
        (4) & \< عشان > & for & \<لكي > &\<تبي بطاريات عشان تلعب > (\textit{she wants batteries to play})  & LOC  \\

       (5) &  \<دلوقتي > & now &  \<الآن >  & \<...اقنعوه ينزل دلوقتي>  \textit{(convince him to move now})  &         PER  \\

       (6) &   \<ايش> & what &  \<ماذا>
          & \<ايش رأيك>
          (\textit{what do you think?}) & PER \\

        (7) & \<قادر > & capable &  \<قادر> &\<وبقدرة قادر...>  (\textit{magically}; idiomatic expression)& PER \\

       (8) &  \<المشين > & shameful &  \<المشين >  & \<المشين طنطاوي >  (\textit{shameful Tantawy}; Playful for \textit{General Tant.})& PER \\

       (9) &  \<ايديكوا > & your hands  &  \<أيديكم >   &   \<ابوس ايديكوا اقنعوه... >   \textit{(I entreat you to convince hi}m) & PER \\

      (10) &  \<اسالك > & I ask you &  \<أسألك> & \< ودي اسالك شنهي > (\textit{I ask you what}) & ORG  \\
      (11) &   \<مين > & who &  \<مين>
          & \<صوتك مع مين البدوي >
          (\textit{who do you vote for, Badawi}) & PER  \\
        
       (12) &   \<فلوبي ديسك> & floppy disk &  \<قرص مرن>  & \<ماسك عليه فلوبي ديسك >  (\textit{holds a floppy disk against him}) & PER  \\

       (13) &   \<لحبايب> & loved ones &  \<الأحباء> & \<تعال علم يف لحبايب > (\textit{come teach your loved ones}) & LOC \\
          
        (14) & \<ماي> & water & \<ماء> & \<جبت لهم ماي> \textit{(brought them water)} & PER  \\

    (15) & \<ريتويت> & retweet & \<إعادة تغريد> & \<لو قرفان دوس  ريتويت> \textit{(if depressed click retweet)} & PER \\
    
   \bottomrule
    \end{tabular}
    \caption{\textbf{NER task.} Bigger sample false positives mitigated by self-training. These were correctly predicted as the unnamed entity ``O" by the self-trained model.}
    \label{tab:app-fp_analysis}
\end{table*}


\begin{table*}[h!]
    \centering
    \footnotesize
     \begin{tabular}{l|c|l|c|c|c}
        \toprule
        \textbf{no.} & \textbf{Token(s)} & \textbf{Context/Explanation} & \textbf{Gold} & \textbf{FT} & \textbf{ST}  \\
        \hline
        (1) & \<بالمبارك> & \<بالمبارك عاد احنا> \textit{(We are still in Mubarak}) & LOC & PER & O \\
        (2) & \<محشش> &   \<محشش دخل المحاضرة>
        \textit{(a drunk entered the lecture)} &  O & PER & PER \\
        (3) & \<بأمارة> & \<بأمارة ايه وفين>
        \textit{(what is the evidence/sign and where?)} &  O & LOC & LOC \\
        (4) & \<لمستفشي> & \<لمستفشي قصر الدوباره>
        \textit{(to Qasr AlDobara Hospital)} & LOC & O & O \\
        (5) & \<كنتاكي> & \<عند كنتاكي>
        \textit{(by Kentucky [resturant])} & LOC & O & O \\
        (6) & \<داون تاون> & \<مشروع داون تاون بطنطا>
        \textit{(a down town Tanta project)}&
        LOC & O & O \\
        (7) & \<يابطل> & \<مبروك يابطل> 
        \textit{(Congratulations, hero!)} &  O & PER & PER \\
        (8) & \<الاخوان> & \<نختلف مع الاخوان>  \textit{(we disagree with the Muslim brotherhood)} & ORG & O & O \\
        (9) & \<قناة العربية> & \<شفت قناة العربية>  \textit{(watched Al Arabya Channel)}&  ORG & O & O \\
        (10) & \<المجلس العسكري> &  \<اللي عمله المجلس العسكري> \textit{(what the military council did)} & ORG & O & O \\
         \bottomrule
    \end{tabular}
    
    \caption{\textbf{NER task.} Sample errors that are not fixed by self-training (shared with the mere fine-tuned model).}
    \label{tab:app-shared-errors}
\end{table*}


\begin{table}[h!]
    \footnotesize 
    \centering

     \begin{tabular}{l|c|c|c|c}
        \toprule
        \textbf{no.} & \textbf{Word}  & \textbf{Gold} & \textbf{FT} & \textbf{ST}  \\
        \hline
        (1) & \<الاخوان > &  ORG  &  ORG  & O \\
        (2) &   \<للبرادعي > &   PER &   PER &  O \\
         (3) &  \< مجدي الجلاد > &  PER &   PER  & O \\
        (4) &   \<فان ديزل>  &  PER &   PER  &  O \\
        (5) &  \<الجنزوري> &  PER &   PER  &  O \\
        (6) &  \<زين يسون>   &  PER &   PER  &  O \\
         \bottomrule
    \end{tabular}
    
    \caption{\textbf{NER task.} Sample false negatives produced by self-training.}
    \label{tab:app-fn-analysis}
\end{table}

\begin{table*}[h!]
    \centering
    \footnotesize
     \begin{tabular}{l|l|c|c}
        \toprule
        \textbf{no} & \textbf{Example} & \textbf{FT} & \textbf{ST}  \\
        \hline
        (1) & \<لن يفهما غير العباقره كنت عامل بيه غساله> & Non-Sarcastic & sarcastic\\
        (2) & \<هذا وضعكم بدون ميسي يا تافهين يا عاهات> & Non-sarcastic & sarcastic \\ 
        (3) & \<حاميها حراميها> & Non-sarcastic & sarcastic \\
        (4) & \<ايز لكل رجل قبيله عايزها متقسمه HASH عبيط رسمي نظمي> & Non-sarcastic & sarcastic \\
        (5) & \<للعلم ده مهرجان درجه اولي ما لو بالفلوس ما كانش حد غلب> & Sarcastic & Non-sarcastic \\ 
        (6) & \<الاستاذه ميريام فارس ليها اغاني حلوه فشخ مش واخده حقها> & Sarcastic & Non-sarcastic  \\
        (7) & \<يا تري بين هيلاري كلنتون ودونالد ترامب مين نختار> & Sarcastic & Non-sarcastic \\
        (8) & \<دعوه الست قصده هيلاري كلنتون مثلا> & Sarcastic & Non-sarcastic \\
         \bottomrule
    \end{tabular}
    
    \caption{\textbf{SRD task.} Sample errors that were fixed by self-training }
    \label{tab:app-SRD-fixed-errors}
\end{table*}

\begin{table*}[h!]
    \centering
    \footnotesize
     \begin{tabular}{l|l|c|c}
        \toprule
        \textbf{no} & \textbf{Example} & \textbf{Prediction} & \textbf{Gold}  \\
        \hline
        (1) & \<عارفصوره وجدتها علي فيسبوك> &sarcastic &non-sarcastic\\
        (2) & \<فضيعه> & sarcastic & non-sarcastic \\ 
        (3) & \<بغنيلا وبدئلا وغير بحبك مابئلا>&  sarcastic & Non-sarcastic \\
        (4) & \<انت الغالي اللي بقالي سنين بهواه> & Sarcastic & Non-sarcastic \\
        (5) & \<هه دنتم مسخره ياراقل هونو علي انفسكم يامجن فرنسا> & non-sarcastic & sarcastic \\
        (6) & \<يعني حيكون زي اللورد دارث فيدرر هه>  &  non-sarcastic & sarcastic \\
        (7) & \<يا جماعه هذا بوكيمون ماحدا عرف يصطاده ويطعميه للجرذان> & non-sarcastic  & sarcastic\\ 
        (8) & \< حضرتك مفيش فكه تاخد بالباقي ريتويتس> & non-sarcastic & sarcastic \\
         \bottomrule
    \end{tabular}
    
    \caption{\textbf{SRD task.} Sample errors that were not fixed by self-training (shared with the mere fine-tuned model).}
    \label{tab:app-SRD-shared-errors}
\end{table*}